\documentclass{sig-alternate-05-2015}

\usepackage{caption} 
\begin{document}

\setcopyright{acmcopyright}

\doi{XX.XXX/XXX_X}

\isbn{XXX-XXXX-XX-XXX/XX/XX}

\conferenceinfo{KDD '16}{August 13--17, 2016, San Francisco, California, USA}

\acmPrice{\$15.00}

%
\conferenceinfo{KDD}{'16 San Francisco, California, USA}

\title{Ranking academic institutions on potential paper acceptance in upcoming conferences}

%
%
%
%
%

\numberofauthors{5} 
%
\author{
%
%
\alignauthor
Jobin Wilson\\
       \affaddr{R\&D Department}\\
       \affaddr{Flytxt, India}\\
       \email{jobin.wilson@flytxt.com}
\alignauthor
Ram Mohan\\
       \affaddr{R\&D Department}\\
       \affaddr{Flytxt, India}\\
       \email{ram.mohan@flytxt.com}
\alignauthor 
Muhammad Arif\\
       \affaddr{R\&D Department}\\
       \affaddr{Flytxt, India}\\
       \email{muhammad.arif@flytxt.com}
\and  
\alignauthor Santanu Chaudhury\\
       \affaddr{Department of EE}\\
       \affaddr{Indian Institute of Technology, Delhi}\\
       \email{santanuc@ee.iitd.ac.in}
\alignauthor
\alignauthor Brejesh Lall\\
       \affaddr{Department of EE}\\
       \affaddr{Indian Institute of Technology, Delhi}\\
       \email{brejesh.lall@ee.iitd.ac.in}
}

\maketitle
\begin{abstract}
The crux of the problem in KDD Cup 2016 involves developing data mining techniques to rank research institutions based on publications. Rank importance of research institutions are derived from predictions on the number of full research papers that would potentially get accepted in upcoming top-tier conferences, utilizing public information on the web.
This paper describes our solution to KDD Cup 2016. We used a two step approach in which we first identify full research papers corresponding to each conference of interest and then train two variants of exponential smoothing models to make predictions. Our solution achieves an overall score of 0.7508, while the winning submission scored 0.7656 in the overall results. 
 
\end{abstract}

%
%


%
%

%
%



\section{Introduction}
\label{sec:introduction}
The goal of KDD Cup 2016 was to build data mining techniques capable of ranking research institutions on the basis of potential paper acceptance in upcoming top-tier conferences, utilizing public data sources available on the web. The concrete task involved ranking institutions by predicting the number of their full research papers that would get accepted in specific upcoming conferences such as SIGIR, KDD and ACM MM,  corresponding to different phases of the contest. 

Predictions corresponding to each conference is a ranked list of institutions on the basis of relevance score (rel score) for a specific upcoming conference in the year 2016. Each accepted full research paper in a particular conference has an equal vote of 1. Each author is considered to have an equal contribution to the paper. In case if an author has multiple affiliations, each affiliation is considered to contribute equally to the paper. We aggregate the fraction of the votes that each affiliations receives from all the full research papers accepted from that institution for a particular conference in a particular year, to calculate its rel score. The rel score thus represents importance of an affiliation in relation to a particular conference in a particular year. As the actual set of accepted papers for an upcoming conference will not be available at the time of prediction, we make use of data mining techniques which utilize historical data points from publicly available data sources on the web to come up with an estimate of the rel score for each institution of interest. 

A major public data source that we made use of was the Microsoft Academic Graph (MAG) \cite{sinha2015overview}, which includes information about publications, citations, authors, affiliations and venues. Since MAG data doesn't differentiate full research papers from others, we also made use of ACM digital library (ACM DL) data corresponding to specific conferences. For each conference of interest, we extracted information about all published papers along with the sections to which they belonged in the proceedings, using a web-crawler script.
\section{Contest Evaluation}
\label{sec:contest_evaluation}
The contest was split into three phases wherein each phase spanned a calender month. Corresponding to each phase, predictions were generated against multiple conferences associated with that phase. At the end of each phase, organizers would select a particular conference from the set and evaluate the predictions by comparing it with the actual list of papers that got accepted by the selected conference. The overall final score was calculated as a weighted sum of the scores from each phase. Phase 1 carried a 20\% weight where as Phase 2 and 3 carried a $40\%$ weight each.

The evaluation metric employed to asses the quality of predictions corresponding to each phase is Normalized discounted cumulative gain (NDCG), a commonly used metric in information retrieval \cite{jarvelin2002cumulated}. NDCG score varies from 0.0 to 1.0, wherein ideal ranking of entities is represented by a value of 1.0. In our problem, NDCG@20 score is calculated as follows.  Once the actual list of accepted full research papers becomes available, rel scores corresponding to each institution gets calculated. Rel score corresponding to the institution at the $i^{th}$ rank position is denoted as $rel_i$.

\begin{equation}
\label{eqn:dcg}
DCG_n=\sum_{i=1}^n \frac{rel_i}{log_2(i+1)}
\end{equation}

\begin{equation}
\label{eqn:ndcg}
NDCG_n=\frac{DCG_n}{IDCG_n}
\end{equation}

DCG@20 is calculated using Equation~\ref{eqn:dcg}. IDCG corresponds to the ideal ranking order of institutions with respect to a conference, calculated from the actual rel scores when the ground truth becomes available. NDCG@20 is calculated from DCG@20 and IDCG@20, as indicated in Equation~\ref{eqn:ndcg}.

\section{Related Work}
There have been several studies which attempt to rank research institutions based on a variety of performance indicators such as published articles in selected top-tier journals,  major international awards and highly cited researchers in prominent fields \cite{liu2005academic}\cite{wu2012ranking}. Though publishing an yearly ranking of research institutions or universities has become a tradition for many academic institutions, newspapers and magazines, quantifying long term scientific impact is still considered a difficult problem \cite{wang2013quantifying}. Diversity in the data considered, methodology employed and subjective aspects involved in the ranking procedures across various ranking approaches impose further challenges to this problem.

A popular ranking methodology called Academic Ranking of World Universities (ARWU) proposed by Liu, Nian Cai, and Ying Cheng utilizes multifarious indicators such as alumni/staff as Nobel laureates and Fields Medalists, highly cited researchers in important fields, articles published in top journals such Nature and Science, and/or indexed by major citation indexes such as Science Citation Index-expanded, and  per capita academic performance \cite{liu2005academic}.

Wu, Hung-Yi, et al. (2012) proposed a hybrid multiple-criteria decision making (MCDM) model for weighing various performance evaluation indices as well as to rank research institutions \cite{wu2012ranking}. Their approach involved using Analytic hierarchy process (AHP) to first weigh various  performance indices belonging to dimensions such as Administration, Teaching Resources, Internationalization, Faculty, Teaching and Research, and subsequently utilizing the weights to generate the ranking of institutions.

Wang, Dashun et al. (2013) proposed an approach to quantify long term scientific impact using a metric called ultimate impact ($c^{\infty}$). This metric provides a journal independent assessment of a paper's long term impact, and it also has a meaningful interpretation  \cite{wang2013quantifying}. $c^{\infty}$ captures the total number of citations a paper will ever acquire.

Due to the unique nature of the KDD Cup 2016 problem and the associated evaluation procedure, we realized that existing approaches for ranking research institutions wouldn't be directly applicable in our context. Our approach is described in detail in Section~\ref{sec:proposed_approach}.

\section{Data Preparation}
\label{sec:data_prep}
We primarily made use of MAG data that was available for download as a part of the KDD Cup 2016, from \cite{website:mag2016}. We also utilized ACM Digital Library (http://dl.acm.org/) which provides information on all the papers that are published in each of the ACM conference proceedings. Along with the paper title, it provides information on the number of pages in a paper as well as the section to which it belongs in the proceedings. Microsoft Academic Knowledge API was also used in the initial phase for validation purpose of the aggregated data.

For data extraction and processing, we downloaded the MAG dataset to a local Apache Spark\cite{zaharia2010spark}  cluster consisting of 4 Nodes, with each node having a configuration of 8 cores, 32 GB RAM and 1TB hard disk. The Spark cluster was running on top of Hadoop with Yarn \cite{vavilapalli2013apache}. Spark SQL framework was used for data pre-processing. The uncompressed data volume corresponding to MAG dataset was around 100 GB. Since rel score calculation for historical instances of conferences would require a join between the files "Papers" and "PaperAuthorAffiliations" with size ~29 GB(aprox. 12.6 million records) and ~18 GB(aprox. 33.7 million records) respectively, details from the "Papers" file corresponding to the 8 conferences of interest was extracted into separate files for optimizing the join.

Identifying full research papers was a problem in itself, and we attempted to solve it in a semi-automated approach utilizing auxiliary data from ACM DL. Our approach is described in detail in Section~\ref{subsec:classification_models}. After extracting the list of full research papers corresponding to a conference of interest, PaperId corresponding to each paper is identified by matching the paper title with the "Normalized paper title" field in the "Papers" file. In case if an exact match is not found, a soft matching procedure using sequence alignment is performed to identify the paper title corresponding to the minimum matching cost, representing the most likely match. Duplicates are also eliminated while performing the title matching.

Once the relevant PaperIds are identified, rel score calculation per institution per year with respect to a conference of interest would only involve a join between "Papers" and "PaperAuthorAffiliations", followed by a simple aggregation procedure described in Section~\ref{sec:introduction}.

\section{Methodology}
\label{sec:methodology}
We attempt to transform the problem of ranking research institutions into a time series forecasting problem. Concretely, relevance score of an affiliation with respect to a conference of interest for a particular year is derived from the list of research papers that got accepted in that conference during that year, using the procedure described in Section~\ref{sec:introduction}. Thus,  for each conference of interest, an ordered sequence of rel scores are extracted corresponding to a research institution, forming a univariate time series. We represent historical rel scores of an institution with respect to a conference as a time series wherein the first observation corresponds to the very first instance of the conference and the last observation corresponds to the most recent instance of the conference. In case if an institution has no accepted papers for a particular conference in a particular year, rel score corresponding to that year would be set as zero in the corresponding time series. We make use of Box-Jenkins models \cite{box2015time} as well as two variants of exponential smoothing models to forecast the rel score of each institution of interest with respect to specific conferences.
We generate the ranked list of research institutions with respect to an upcoming conference of interest by ordering affiliations on the basis of predicted rel scores in the descending order.

As our main ranking problem got transformed into a time series forecasting problem, time series modeling was one approach that we strongly pursued. We also explored causal modeling using Bayesian networks as well as graph based ranking methods such as Page Rank on co-authorship networks, to address the research institution ranking problem.

An auxiliary problem that we had to solve was to identify full research papers accepted by a conference of interest from the set of all accepted papers. We attempted to solve this problem in a semi-automated fashion using classification models along with manual data analysis.

Since our submissions in each phase of the contest made use of different models, we will first introduce them along with some of our experimental observations. In Section~\ref{sec:proposed_approach}, we provide a detailed treatment of the procedures that we followed for generating our actual submissions across phases. 

\subsection{Time Series Models}

\subsubsection{ARIMA Model}
Autoregressive Integrated Moving average (ARIMA) model is a generalization of an autoregressive moving average model
(ARMA) and is applied when the original data appears non-stationary \cite{box2015time}. The initial differencing step 
is applied to reduce non-stationarity. ARMA model is constituted by a combination of two models namely Autoregressive (AR) and Moving average(MA). In the case of AR model represented by AR(p), the future value of a variable is expressed as a linear combination of the $p$ past observed values together with an error term and a constant. Equation~\ref{eqn:ar} represents an AR(p) model where $y_t$ is the observed value and $\epsilon_t$ is the random error observed at time $t$ \cite{box2015time}. Parameters of the model are represented by the values $\phi_i$ for $i=1,2...p$ and the constant $c$. Order of the model is denoted by $p$.

\begin{equation}
\label{eqn:ar}
y_t=\sum_{i=1}^{p}\phi_iy_{t-i}+\epsilon_t+c
\end{equation}

In the case of an MA(q) model, the future value of a variable is expressed as a linear combination of past error values and a mean value of the observed variable. Equation~\ref{eqn:ma} represents an MA(q) model where $\mu$ denotes the mean values of the sequence, $\theta_j$ where $j=1,2...q$ represents the model parameters and $q$ denotes the order of the model.

\begin{equation}
\label{eqn:ma}
y_t= \mu + \sum_{j=1}^{q}\theta_j\epsilon_{t-j} + \epsilon_t
\end{equation}

ARMA(p,q) models \cite{box2015time} combine AR(p) and MA(q) models where $p$ and $q$ are the model orders corresponding to $p$ autoregressive terms and $q$ moving average terms, as represented in the Equation~\ref{eqn:arma} 

\begin{equation}
\label{eqn:arma}
y_t=\sum_{i=1}^{p}\phi_iy_{t-i}+\epsilon_t+c + \sum_{j=1}^{q}\theta_j\epsilon_{t-j}
\end{equation}

We observe that length of the rel score time series vary significantly across conferences. For instance, proceedings of older conferences such as SIGMOD and SIGIR start from 1970 and 1971 respectively, resulting in
relatively longer time series having 44-45 units of length were as conferences such as KDD have a relatively shorter history, resulting in time series which are only 21 units long. Publication history and consistency of research institutions also varied significantly. For instance, relatively older institutions such as IBM had a longer and more consistent history with respect to certain conferences of interest, compared to relatively newer ones such as facebook. Based on these observations as well as the fact that rel score time series may not always be stationary to fit an ARIMA model, we explore few simple time series models also.

\subsubsection{Exponential Smoothing}
Exponential smoothing can be thought of as a special case of weighted average wherein all data points are considered, while assigning exponentially smaller weights as we go back in time. Thus, exponential smoothing continuously refines its forecast on the basis of most recent observations. We make use of single exponential smoothing, which is popularly used for short-range forecasting. The model assumes that the observed data fluctuates around a reasonably stable mean, and doesn't have a consistent pattern of growth or trend \cite{kalekar2004time}. Equation~\ref{eqn:exp_smooth} denotes the formula for simple exponential smoothing.

\begin{equation}
\label{eqn:exp_smooth}
\hat{y}_t=\alpha y_t+(1-\alpha)\hat{y}_{t-1}
\end{equation}

The recursive definition indicates that each new forecast (smoothed value) is a weighted average of the current observation and the previous forecast. The previous forecast in turn was computed from the previous observed value and the forecasted value before the previous observation and so on. The model parameter $\alpha$ is called smoothing factor or smoothing coefficient. The value of $\alpha$ determines how much weight should be given to the most recent observed value  versus the last forecasted.

We observe that determining the correct value of $\alpha$ would be crucial for the model's predictive power. Since the nature of the conferences of interest varied significantly, the importance of the most recent rel score of an institution with respect to a conference in determining its position in the upcoming instance also vary. Based on this fact, we empirically determine optimal $\alpha$ corresponding to each conference of interest using a grid-search along with a cross-validation procedure. Details of the procedure is described in Section~\ref{sec:phase3}.

\subsubsection{Naive Exponential Smoothing}
Inspired from the key ideas of exponential smoothing and based on our intuition that recent ranking of a research institution with respect to a conference is likely to be more indicative of its rank position in the upcoming instance of that conference, we formulated a simple non-parametric exponential smoothing scheme as represented in Equation~\ref{eqn:naive_exp_smooth}.

\begin{equation}
\label{eqn:naive_exp_smooth}
y_{t+1} = \sum_{i=1}^{t}\frac{y_i}{e^{t-i}}
\end{equation}

In this model, forecast for the next instance of time is a weighted sum of all the previously observed values with an exponentially decreasing weight as we go back in time.

Despite of its simplicity, we observe that this model works quite well for our institution ranking problem and has the advantage that it is non-parametric, making prediction an easy task.

\subsection{Causal Model}
Causal modeling using Bayesian networks is one approach in which we attempt to learn how various factors such as count of authors in an institution within a particular H-index range, count of authors in an institution with a particular publication frequency range for a conference of interest, weather the institution is academic or industry etc. influence the institutions publication in an upcoming conference of interest.

The causal model is represented as a directed acyclic graph with joint probability distribution factorizing according to the graph. Graphical structure and conditional independencies are used to capture the structure of our ranking problem. Learning task from historical data is performed in two steps. Causal structure learning is performed using ParallelPC \cite{hauser2012characterization} package and parameter learning is performed using the bnlearn\cite{scutari2009learning} package; both available as part of the RStudio. Generating predictions involve calculating the expected rel scores of institutions with respect to an upcoming conference of interest, after performing belief propagation on the learned network. Though the model's cross-validation scores were impressive for certain years, high variance in the scores suggested a potential generalization issue.

\subsection{Network Model}
We experimented with a simple network model by constructing an affiliation network based on co-authorship wherein edge weights indicate the number of times co-authorship occurred among the institutions connected by that edge, with respect to a particular conference of interest. Our assumption was that prominence a nodes within this network could potentially indicate its ranking. Page Rank algorithm was used to generate affiliation rankings from this graph. We observed that the cross-validation scores for this model was low.

\subsection{Classification Models}
\label{subsec:classification_models}
In order to identify full research papers accepted in a particular conference in a particular year, we started off with a manual process which proved to be very time consuming. We manually downloaded webpages with the list of accepted full research papers from the conference website corresponding to each year and extracted the paper names by parsing the html content using a python script. We improved this procedure by implementing a web-crawler script which automatically downloaded the conference proceedings corresponding to each instance of the conference of interest from the ACM Digital Library (http://dl.acm.org/).  To determine if a paper was a full research paper, we initially used manual rules based on the section names in the proceeding to which it belonged. Based on the intuition that total number of pages in a paper and the name of the section to which it belonged in the proceedings could be potentially used to identify full research papers, we formulated a two-class classification problem to predict if a paper was a full research paper or not, in Phase 3. Labelled training data was generated from the set of full research papers that was provided along with MAG data, after a manual verification step.

\subsubsection{SVM}
Support Vector Machines (SVMs) are a set of supervised learning techniques suitable for classification, regression and outlier detection. SVMs are based on Structural Risk Minimization principle by Vapnik(1995) from computational learning theory \cite{joachims1998text}\cite{hearst1998support}. There have been several studies which argue that SVMs are well suited for text categorization tasks and show substantial performance gains compared with many popular models \cite{joachims1998text}.

We make use of the LinearSVC implementation of SVM classifier in scikit-learn package\cite{pedregosa2011scikit}. We extract  features corresponding to the section names in the proceedings and make use of tf-idf based vectorization scheme to generate feature vectors to train a classifier. The trained classifier takes in a section name within the conference proceedings and predict if that section is likely a full-research paper section or not.

\subsubsection{Random Forests}
Random Forests is a tree based ensemble learning method proposed by Leo Breiman \cite{breiman2001random}. An ensemble consisting of several decision trees is constructed during the model training phase by sub-sampling features randomly. The model produces output by averaging the predictions of individual decision trees within the forest. This approach improves generalization and reduces the prediction variance and the effect of noise.

We make use of the RandomForestClassifier implementation within scikit-learn package. Based on our intuition that full research papers are unlikely to be very short, we used a single feature namely paper length, the total number of pages in a research paper, to train a Random Forest model.

\section{Our Approach}
\label{sec:proposed_approach}
In this section, we present the specific procedures that were followed to generate our submissions corresponding to each phase, by utilizing one or more of the models described earlier. Cross-validation was applied to determine the quality of predictions using the NDCG metric described in Section~\ref{sec:contest_evaluation}. We implemented a cross-validation framework that took a prediction file, conference Id, prediction year and a parameter $N$ as inputs and generated the corresponding NDCG@N score by comparing predicted rankings with the actual rankings. To forecast an institution's rel score for the $t^{th}$ instance of a conference, our models utilize the institution's rel scores from the first instance till the $t-1^{th}$ instance of that conference. In each phase, an expected value of the NDCG@20 score corresponding to our predictions for an upcoming conference of interest was estimated by averaging the corss-validation NDCG scores corresponding to the previous 3 instances of that conference.

\subsection{Phase 1}
Our phase 1 approach rely on MAG data alone for extracting rel scores corresponding to research institutions across years. Since MAG doesn't differentiate full research papers from others, we assume that rel scores extracted by considering all accepted papers from a research institution in a conference of interest would be a good indicator of its ranking in the upcoming instance of that conference. Phase 1 results disproved our assumption as our actual score was significantly below our average cross-validation NDCG score.

Our ranking procedure for this phase involves learning an individual ARIMA model corresponding to each research institution using its yearly rel score time series for a conference of interest followed by forecasting its rel score in the upcoming instance of that conference. We make use of the tsa.ARIMA model within the statsmodels package\cite{seabold2010statsmodels}. The order of the model is specified as a tuple $(p,d,q)$, where $p,d,q$ denotes the number of AR parameters, differences and MA parameters respectively. Corresponding to each rel score time series, a set of 3 tuples (1,1,1), (1,1,0) and (0,1,1) are used as candidates to train the model. The tuple which minimize the Root Mean Squared Error (RMSE) in a cross-validation setting is selected to train the final ARIMA model corresponding to each research institution. In case if the rel score time series corresponding to an institution is non-stationary for a conference of interest, average rel score of the institution in the last 3 instances of that conference becomes the forecasted rel score for the upcoming instance.

Based on this approach, we observe that our average cross-validation NDCG score for SIGIR was 0.8694 where as our score in the actual Phase 1 result was only 0.6721, taking us to the rank 104.

In retrospect, we recalculate the rel scores by considering only the accepted full research papers in SIGIR over the years, and fit an ARIMA(1,1,1) model for each institution to generate predictions. Full research papers were identified manually in this phase. Non-stationary time series were handled in the same way as we explained earlier. The results are captured in Table 1. We observe that ARIMA(1,1,1) is a promising model for SIGIR when trained with the correct rel scores, as the average cross-validation NDCG@20 score comes to 0.8323 whereas the winning submission score for Phase 1 is 0.8273. 

\begin{table}
\label{tab:phase_1_retrospect}
\centering
\caption{SIGIR NDCG cross-validation scores}
\begin{tabular}{|c|c|c|c|} \hline
\textbf{Year} & \textbf{NDCG@20} & \textbf{NDCG@100} & \textbf{NDCG@150}\\ \hline
2015 & 0.7761 & 0.7587 & 0.7675\\ \hline
2014 & 0.8777 & 0.7911 & 0.7978\\ \hline
2013 & 0.8432 & 0.7766 & 0.7894\\ \hline
Avg & 0.8323 & 0.7755 & 0.7849\\ \hline
\end{tabular}
\end{table}

\subsection{Phase 2}
In Phase 2, full research papers are identified by utilizing ACM DL conference proceedings data as an auxiliary source. We extract section name in the proceedings to which a paper belongs as well as the length of the paper from the conference proceedings using a parser. ICML proceedings corresponding to certain years were unavailable in ACM DL. We manually collect the proceedings corresponding to those years. 

While parsing the proceedings, full research papers are identified using simple rules based on section names. For instance, sections having names such as "Keynote", "Panel", "Industry Track", "Posters" etc. are less likely to have full research papers within them. We construct filtering rules by manually adding such keywords into a dictionary and then use substring matching to decide if a section is likely to contain full research papers.

After extracting the list of full research papers accepted by a conference of interest across years, we generate the corresponding rel scores as described in Section~\ref{sec:data_prep}. Cross-validation is used to identify the best model corresponding to each conference of interest.  Our cross-validation results for KDD are captured in Table 2.

\begin{table}
\label{tab:KDD_crossvalidation}
\centering
\caption{KDD NDCG cross-validation scores}
\begin{tabular}{|c|c|c|} \hline
\textbf{Model} & \textbf{Year} & \textbf{NDCG@20} \\ \hline
Naive Exponential Smoothing & 2015 & 0.8110\\ \hline
Naive Exponential Smoothing & 2014 & 0.7411\\ \hline
Naive Exponential Smoothing & 2013 & 0.8342\\ \hline
ARIMA(1,1,1) & 2015 & 0.7249\\ \hline
ARIMA(0,1,1) & 2015 & 0.7250\\ \hline
\end{tabular}
\end{table}

For KDD, we observe that Naive Exponential Smoothing performs significantly better than ARIMA models and provide an average NDCG@20 score of 0.7954 during cross-validation. Phase 2 result confirms that our model generalizes well, as our score in the actual result was 0.8075, taking us to the 2nd rank position in that phase.

\subsection{Phase 3}
\label{sec:phase3}
We observe that Phase 3 is significantly complicated  compared to the previous phases, as the top performing institutions are not consistent across time for the conferences of interest. For instance, for the conference ACM MM, the number of unique affiliations that ever came to top 20 positions in the last 5 years (2011-2015) were 67. 

For identifying full research papers, we follow the procedure described in Section~\ref{subsec:classification_models}. Though manual rules based on paper length alone could potentially provide an approximate list of full research papers, we utilize classification models and manually review their predictions to minimize false positives. 

Poor performance of ARIMA models is attributed to high fluctuations in the rel score data corresponding to this phase. During cross-validation, we observe that naive exponential smoothing and exponential smoothing models perform comparatively better. We perform a grid-search to tune the smoothing parameter $\alpha$ for the exponential smoothing model. The optimal values for $\alpha$ corresponding to the conferences ACM MM, FSE and MobiComm is empirically determined as 0.4, 0.2 and 0.35 respectively. Our cross-validation results are captured in Table 3.

\begin{table}
\label{tab:MM_crossvalidation}
\centering
\caption{ACM MM NDCG cross-validation scores}
\begin{tabular}{|c|c|c|} \hline
\textbf{Model} & \textbf{Year} & \textbf{NDCG@20} \\ \hline
ARIMA(1,1,1) & 2015 & 0.5546\\ \hline
Naive Exponential Smoothing & 2015 & 0.6487\\ \hline
Naive Exponential Smoothing & 2014 & 0.7343\\ \hline
Naive Exponential Smoothing & 2013 & 0.7263\\ \hline
Exponential Smoothing & 2015 & 0.6405\\ \hline
Exponential Smoothing & 2014 & 0.8086\\ \hline
Exponential Smoothing & 2013 & 0.7391\\ \hline
\end{tabular}
\end{table}

We select exponential smoothing model with $\alpha=0.4$ to generate predictions for ACM MM, as the average cross-validation NDCG@20 score is 0.7294. Our actual score in the Phase 3 result was 0.7334, taking us to the  51th rank position in that phase.

\section{Discussion}
We observe that KDD Cup 2016 was unique and challenging in several ways. The open nature of the problem, non-availability of ground truth data beforehand, several possible solution approaches based on diverse facets of academic ranking such as citation networks, co-authorship networks, similarity among conferences, conference locations, temporal information etc. all added to the difficulty of the challenge.

Getting to the right data was one of the most important challenge that we faced. Concretely, identifying full research papers from the set of all accepted papers in a conference was non-trivial. Our semi-automated approach for this task by considering ACM DL as an auxiliary data source was in fact a natural progression from a completely manual procedure to a semi-automated mechanism involving a web-crawler and multiple classification models, emphasizing the importance of human-in-the-loop machine learning pipelines \cite{ware2001interactive}. Also, powerful distributed computing frameworks such as Apache Spark and Hadoop helped us to easily deal with the large data volume of MAG. Spark SQL simplified our data pre-processing tasks as many of them were expressible as simple SQL queries.

Model selection was yet another important aspect. Since the nature of each conference of interest significantly varied,  the corresponding rel score time series also had diverse characteristics. This made it essential to have a strong cross-validation framework for selecting the most appropriate model and its parameters for each of the conference of interest. Since our cross-validation framework used NDCG@20 itself as the evaluation metric, we were able to derive a reasonable estimate of our actual result scores, by averaging our cross-validation scores.

Finally, we observed that once the right data is extracted, even simple models such as exponential smoothing provides good predictive performance and is able to generalize well in this case.
\section{Conclusion}
In this paper, we presented our solution to KDD Cup 2016. We transformed the original research institution ranking problem into a time series forecasting problem. Our solution involved a two step procedure in which we first identified full research papers corresponding to a conference of interest using a semi-automated procedure and then applied time series models such ARIMA and exponential smoothing to generate predictions. We implemented a cross-validation framework for selecting the most appropriate model and its parameters corresponding to each conference of interest. This framework proved to be an essential tool that helped us to gauge how well our models generalized. Our solution achieved an overall score of 0.7508, whereas the winning submission scored 0.7656 in the overall results.
%
\bibliographystyle{abbrv}
\bibliography{sigproc}  
%
%
\end{document}